\documentclass{article}
\usepackage[square,numbers]{natbib}

\usepackage[preprint]{neurips_2024}

\usepackage[utf8]{inputenc} 
\usepackage[T1]{fontenc}    
\usepackage{hyperref}       
\usepackage{url}            
\usepackage{booktabs}       
\usepackage{amsfonts}       
\usepackage{nicefrac}       
\usepackage{microtype}      
\usepackage{xcolor}         
\usepackage{amsmath}
\usepackage[pdftex]{graphicx}
\title{GAN-TAT: A Novel Framework Using Protein Interaction Networks in Druggable Gene Identification}

\author{
  George Y. Wang\\
  Oak Park High School\\
  Oak Park, CA 91377\\
  \texttt{george.yuanji.wang@gmail.com}\\
  \And
  Srisharan Murugesan\\
  Oak Park High School\\
  Oak Park, CA 91377\\
  \texttt{srish.murgesh@gmail.com}\\
  \AND
  Aditya P. Rohatgi\\
  Oak Park High School\\
  Oak Park, CA 91377\\
  \texttt{aditya.prince.rohatgi@gmail.com} \\
}

\begin{document}

\maketitle

\begin{abstract}
   Identifying druggable genes is essential for developing effective pharmaceuticals. With the availability of extensive, high-quality data, computational methods have become a significant asset. Protein Interaction Network (PIN) is valuable but challenging to implement due to its high dimensionality and sparsity. Previous methods relied on indirect integration, leading to resolution loss. This study proposes GAN-TAT, a framework utilizing an advanced graph embedding technology, ImGAGN, to directly integrate PIN for druggable gene inference work. Tested on three Pharos datasets, GAN-TAT achieved the highest AUC-ROC score of 0.951 on Tclin. Further evaluation shows that GAN-TAT's predictions are supported by clinical evidence, highlighting its potential practical applications in pharmacogenomics. This research represents a methodological attempt with the direct utilization of PIN, expanding potential new solutions for developing drug targets. The source code of GAN-TAT is available at \url{https://github.com/george-yuanji-wang/GAN-TAT}.

\end{abstract}

\section{Introduction}

The Illuminating the Druggable Genome (IDG) project has highlighted the urgent need to expand our repertoire of druggable genes, which are genes encoding proteins that can be targeted for therapeutic effects \cite{Illuminating-2024-06-26}. Despite significant efforts, many challenges persist \cite{Chakraborty2024}. Therefore, there is a critical need for a systematic and innovative approach to thoroughly investigate and identify these genes, which is essential for advancing pharmaceutical research \cite{Finan_2017}.

The growing availability of extensive multi-omics and systems biology data has provided researchers with unprecedented opportunities to identify druggable genes \cite{Rakshit_2023}. Artificial intelligence (AI) has become instrumental in this process due to its capacity to analyze multidimensional and large-scale biomedical data \cite{Reel_2021, Yu_2022}. For instance, DrugMiner developed a druggability model using protein sequence features, while DrugnomeAI advanced this field by incorporating a heterogeneous set of features, including druggability, generic, and disease-specific attributes, to train ensemble models \cite{Jamali_2016, Raies_2022}.

Protein Interaction Network (PIN) offers a detailed and comprehensive view of protein interactions within biological systems, making them valuable for identifying potential targets \cite{Hao_2016, Rasti_2018}. However, the high dimensionality and vast scale of PINs pose significant challenges for direct computational analysis. Consequently, researchers often resort to indirect approaches, employing summarizing features like topological metrics (e.g., indegree, betweenness, clustering coefficient) to represent these networks \cite{Chua_2014,Wang_2019}. While practical, this utilization leads to a substantial loss of resolution, potentially overlooking subtle but crucial network characteristics essential for identifying druggable genes.

To address this issue, network embedding techniques can be employed \cite{Yan_2007}. Network embedding aims to maximally preserve a network's information while reducing its dimensionality, facilitating higher resolution and better quality of network data \cite{PengCuiUnknown}. Shingo Tsuji successfully used a deep neural network (DNN) to embed a PIN into latent space, creating a framework for inferring Alzheimer’s disease targets \cite{Tsuji_2021}. Building on this, more advanced AI-powered embedding algorithms, such as those based on Random Walk, Graph Neural Networks (GNNs), and edge sampling, offer promising improvements for leveraging PIN data to identify druggable genes \cite{Huang_2021,CanM.LeUnknown,Zhou2018}.

This study aims to address existing constraints in the utilization of the Protein Interaction Network (PIN) for identifying druggable genes. We propose a novel framework, GAN-TAT (Generative Adversarial Network-based Target Assessment Tool), which incorporates a latent representation of the PIN for each gene, serving as a unique feature in a machine-learning model. This representation is generated by the ImGAGN algorithm (Imbalanced Network Embedding via Generative Adversarial Graph Networks), specifically designed to tackle the challenge of imbalance \cite{Qu2021}. Comparative analyses show that GAN-TAT outperforms architectures based on traditional embedding models in efficacy metrics. Validations using three distinct label sets from the Pharos database confirm its effectiveness. Further analysis demonstrates that the gene predictions made by GAN-TAT strongly correlate with clinically validated targets, underscoring its potential as a robust methodological approach for future research and practical applications in pharmacogenomics. 


\section{Method \& Materials}

\subsection{PIN}
\label{PIN}

This study examined a curated set of 6,048 genes that are involved in the cell signaling process. The Protein Interaction Network (PIN) was constructed using signal transduction pathway data from SignaLink 3.0, comprising 6,048 nodes and 20,697 directed, unweighted edges \cite{Csabai_2021}. Additionally, 20 UniProt annotations corresponding to the proteins encoded by these genes were integrated as node attributes \cite{2016}.

\subsection{Extended Feature Set}
\label{Extended_feature_set}

To enhance the performance of downstream classification, we created an extended feature set. Binary-encoded pathway interaction data for each gene was sourced from the Comparative Toxicogenomics Database (CTD) \cite{Davis_2022}. Similarly, chemical interaction data from the CTD was used to create binary features representing the total interactions per gene for different interaction types \cite{Davis_2022}. Gene classification data, including domains, families, and superfamilies from InterPro, were also converted into binary features \cite{Hunter_2009}. The Drug-Gene Interaction Database (DGIDB) provided data on the frequency of gene interactions with various drug types \cite{Cannon_2023}. Additionally, genomic attributes, such as expression levels and variants, were extracted from gnomAD \cite{Karczewski_2020}. Features exhibiting low occurrence frequencies were manually excluded. A Pearson correlation coefficient threshold of 0.85 was subsequently applied, ultimately yielding a refined feature set comprising 324 elements \cite{Schober_2018}.

\subsection{Architecture}
GAN-TAT comprises two primary components: an upstream embedding module and a downstream classification module. The upstream module utilizes the ImGAGN-GraphSAGE model for network embedding, which is a supervised learning model independently trained and optimized using the PIN and a label set. After training, the model generates an 80-dimensional embedding for each gene. These embeddings are then concatenated with the extended feature set to form a comprehensive feature vector, which is subsequently used in the downstream classification module.

\begin{figure}
  \centering
  \fbox{\includegraphics[width=0.975\linewidth]{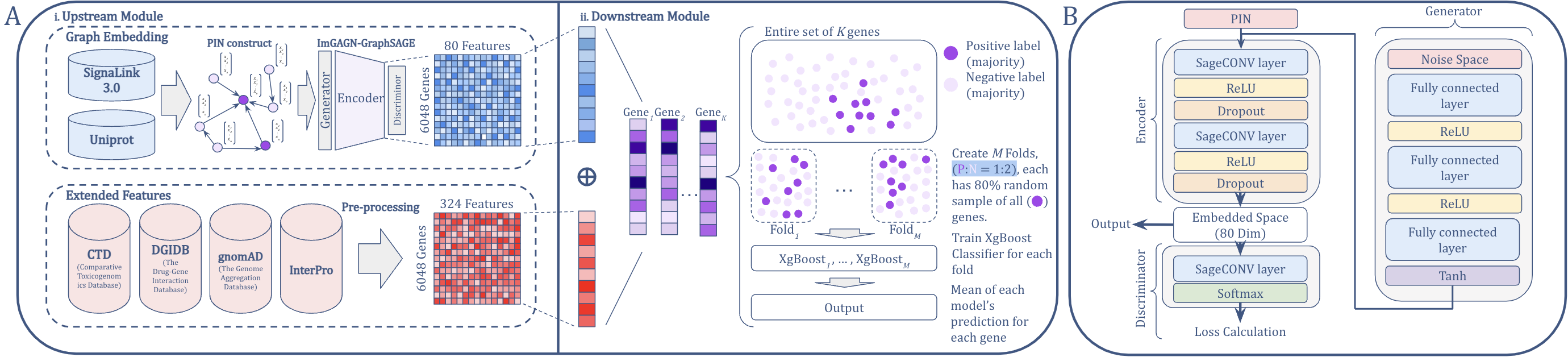}} 
  \caption{\textbf{A)} Illustration of the GAN-TAT architecture. The upstream module embeds the PIN and generates an extended feature set. The downstream module partitions the dataset and trains classifiers. \textbf{B)} Designs of ImGAGN-GraphSAGE, with graph generator, encoder, and discriminator.}
  \label{fig1}
\end{figure}

In the downstream module, a subsampling strategy is implemented to mitigate class imbalance. The complete dataset is partitioned into \textit{M} folds. For each fold, \(80\%\) of the minority (positive) class data points are randomly selected. Additionally, a subset of the majority (negative) class data points is randomly selected to maintain a 1:2 ratio of positive to negative samples. An XgBoost classifier is then trained on each fold. The final prediction score (probability) for each gene is derived by averaging the classifier predictions across all folds \cite{Chen_2016}. Figure~\ref{fig1}A described the architecture.

\subsection{ImGAGN-GraphSAGE}

The ImGAGN model integrates a generative adversarial network (GAN) framework consisting of three principal components: a graph generator, an encoder, and a discriminator \cite{Goodfellow2014,Qu2021}. This architecture effectively manages imbalanced and sparse networks, making it ideal for this study. The generator creates synthetic nodes and edges to achieve a balanced positive-to-negative label ratio of 1:1. After integrating the generated data, the encoder (typically a GNN), produces node embeddings. The discriminator differentiates between genuine and synthetic nodes and classifies their types based on these embeddings. The discriminator is trained for 20 epochs for each epoch of generator training. In this study, the generator comprises three fully connected layers with ReLU activations, culminating in a Tanh function \cite{XavierGlorotUnknown}. The encoder uses the GraphSAGE architecture, with two SAGEConv layers separated by ReLU activations and dropout regularization \cite{XavierGlorotUnknown,Hamilton2017, NitishSrivastavaUnknown} (Figure~\ref{fig1}B). This encoder embeds the graph into an 80-dimensional latent space. The discriminator employs a single SAGEConv layer for classification, concluding with a softmax function. The design is illustrated in Figure~\ref{fig1}B.

\section{Experiment}

The efficacy of various GAN-TAT configurations and frameworks, based on different embedding algorithms, was evaluated using three distinct label sets sourced from Pharos: Tclin genes, Tclin targets for pancreatic intraductal papillary-mucinous neoplasm, and Tclin targets for acute myeloid leukemia\cite{Kelleher_2022}. The PIN and the extended feature set used are the same as described in Sections \ref{PIN} and \ref{Extended_feature_set}.

We trained nine models using three different embedding algorithms: Node2Vec, LINE (Large-scale Information Network Embedding) \cite{Grover2016,Tang_2015}, and ImGAGN-GraphSAGE. Each embedding algorithm was paired with three different classifiers: Decision Tree (DT), Random Forest (RF), and XgBoost\cite{Rokach,Breiman_2001,Chen_2016}. All models were trained on an Apple M3 Pro CPU. Hyperparameters tuning for the ImGAGN-GraphSAGE model was done manually. Node2Vec and LINE were paired with the XgBoost classifier for evaluation. Both these embeddings and all classifiers were tuned using grid search with 5-fold cross-validation and evaluated using the mean AUC-ROC score \cite{Myung_2013, KarimollahHajian-TilakiUnknown}. Further details on hyperparameters and training are available in the GitHub repository. Results are summarized in Table~\ref{performance-table}.

To further assess GAN-TAT's applicability, we grouped genes by their probability percentiles as assigned by the model and mapped them to Tclin genes to calculate overlaps (Figure ~\ref{fig2}A). A Fisher's exact test was performed on these overlaps. Additionally, we compared pathway enrichment scores of the top \(5\%\) predicted genes for GO:BP pathways against those of Tclin genes (Figure ~\ref{fig2}B) \cite{Hill_2008}. A 15-point moving average is applied to reduce noise \cite{Hyndman_2011}.

\section{Result}

Our observations indicate that models based on the ImGAGN-GraphSAGE framework consistently outperform those utilizing other embedding algorithms, particularly in label sets characterized by higher imbalance. Among the classifiers evaluated, XgBoost emerged as the most effective across all embedding methods. The data shows that the GAN-TAT configuration used in this study (ImGAGN-GraphSAGE + XgBoost) stands out as the most superior, with high AUC-ROC scores of 0.951, 0.919, and 0.925 across the datasets for Tclin, pancreatic neoplasm, and leukemia respectively (Table~\ref{performance-table}).

\begin{table}
  \caption{Mean AUC score of different model architecture on three Pharos datasets}
  \label{performance-table}
  \centering
  \begin{tabular}{llccc}
    \toprule
    \multicolumn{2}{c}{Model} & \multicolumn{3}{c}{AUC-ROC} \\
    \cmidrule(lr){1-2} \cmidrule(lr){3-5}
    Embedder & Classifier & Tclin & Pancreatic & Leukemia \\
    \midrule
    Node2Vec & DT & 0.905 & 0.853 & 0.885 \\
    & RF & 0.944 & 0.894 & 0.874 \\
    & XgBoost & 0.940 & 0.892 & 0.914 \\
    \midrule
    LINE & DT & 0.910 & 0.848 & 0.904 \\
    & RF & 0.936 & 0.880 & 0.866 \\
    & XgBoost & 0.947 & 0.899 & 0.929 \\
    \midrule
    ImGAGN-GraphSAGE & DT & 0.950 & \textbf{0.925} & 0.904 \\
    & RF & 0.943 & 0.898 & 0.915 \\
    (GAN-TAT) & XgBoost & \textbf{0.951} & 0.919 & \textbf{0.931} \\
    \bottomrule
  \end{tabular}
\end{table}

\begin{figure}
  \centering
  \fbox{\includegraphics[width=0.975\linewidth]{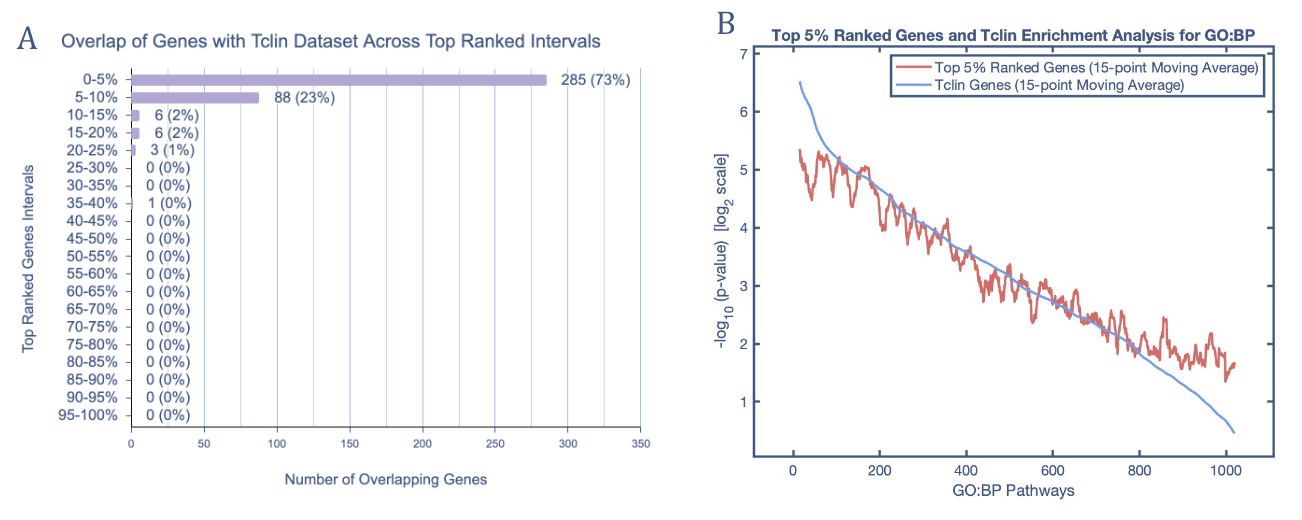}} 
  \caption{\textbf{A)} A bar graph representing the overlap between Tclin drug genes and top predictions of GAN-TAT. \textbf{B)} Comparison of enrichment analysis between top \(5\%\) ranked genes and Tclin.}
  \label{fig2}
\end{figure}

Fisher's exact test indicated extreme enrichment of Tclin genes in the top \(5\%\) interval with a \( \text{p-value} < 2.4 \times 10^{-266}\), and significant enrichment in the \(5-10\%\) interval with a \( \text{p-value} < 1.5 \times 10^{-17}\) \cite{Kim_2017}. The pathway enrichment analysis revealed that GAN-TAT's top \(5\%\) predictions closely align with Tclin genes, as indicated by the two nearly overlapping lines in Figure~\ref{fig2}B. These results suggested that genes predicted by GAN-TAT are supported by clinical evidence.

\section{Conclusion}

This study introduces GAN-TAT, a novel machine-learning framework that leverages network embedding for identifying druggable genes. Utilizing the ImGAGN structure, GAN-TAT helps mitigate challenges associated with the use of PIN. The results show the promising value of GAN-TAT, both through computational performance and validation against clinical data. However, inherent limitations of PIN embedding, such as imbalance, sparsity, and potential overfitting, could effect GAN-TAT's efficacy. Despite these challenges, this work provided an innovative solution to this topic and offers a promising direction for continued research and enhancement.

\newpage
\bibliography{bibliography_citedrive}

\begin{thebibliography}{10}

\bibitem{2016}
Uniprot: the universal protein knowledgebase.
\newblock {\em Nucleic Acids Research}, 45(D1):D158–D169, nov 2016.

\bibitem{Illuminating-2024-06-26}
Illuminating the druggable genome (idg) | nih common fund.
\newblock \url{https://commonfund.nih.gov/idg}, 2024.
\newblock Accessed: 2024-06-26.

\bibitem{Breiman_2001}
Leo Breiman.
\newblock Random forests.
\newblock {\em Machine Learning}, 45(1):5–32, 2001.

\bibitem{Cannon_2023}
Matthew Cannon, James Stevenson, Kathryn Stahl, Rohit Basu, Adam Coffman, Susanna Kiwala, Joshua F McMichael, Kori Kuzma, Dorian Morrissey, Kelsy Cotto, Elaine R Mardis, Obi L Griffith, Malachi Griffith, and Alex H Wagner.
\newblock Dgidb 5.0: rebuilding the drug–gene interaction database for precision medicine and drug discovery platforms.
\newblock {\em Nucleic Acids Research}, 52(D1):D1227–D1235, nov 2023.

\bibitem{Chakraborty2024}
Sohini Chakraborty and Satarupa Banerjee.
\newblock {\em Systems Approaches in Identifying Disease-Related Genes and Drug Targets}, pages 195--255.
\newblock Springer Nature Singapore, Singapore, 2024.

\bibitem{Chen_2016}
Tianqi Chen and Carlos Guestrin.
\newblock Xgboost: A scalable tree boosting system.
\newblock In {\em Proceedings of the 22nd ACM SIGKDD International Conference on Knowledge Discovery and Data Mining}, KDD ’16. ACM, aug 2016.

\bibitem{Chua_2014}
Huey~Eng Chua, Sourav~S. Bhowmick, and Lisa Tucker-Kellogg.
\newblock One feature doesn’t fit all: characterizing topological features of targets in signaling networks.
\newblock In {\em Proceedings of the 5th ACM Conference on Bioinformatics, Computational Biology, and Health Informatics}, BCB ’14. ACM, September 2014.

\bibitem{Csabai_2021}
Luca Csabai, Dávid Fazekas, Tamás Kadlecsik, Máté Szalay-Bekő, Balázs Bohár, et~al.
\newblock Signalink3: a multi-layered resource to uncover tissue-specific signaling networks.
\newblock {\em Nucleic Acids Research}, 50(D1):D701–D709, oct 2021.

\bibitem{PengCuiUnknown}
Peng Cui, Xiao Wang, J.~Pei, and Wenwu Zhu.
\newblock A survey on network embedding.
\newblock {\em IEEE Transactions on Knowledge and Data Engineering}.

\bibitem{Davis_2022}
Allan~Peter Davis, Thomas~C Wiegers, Robin~J Johnson, Daniela Sciaky, Jolene Wiegers, and Carolyn J Mattingly.
\newblock Comparative toxicogenomics database (ctd): update 2023.
\newblock {\em Nucleic Acids Research}, 51(D1):D1257–D1262, sep 2022.

\bibitem{Finan_2017}
Chris Finan, Anna Gaulton, Felix~A. Kruger, R.~Thomas Lumbers, Tina Shah, Jorgen Engmann, Luana Galver, Ryan Kelley, Anneli Karlsson, Rita Santos, John~P. Overington, Aroon~D. Hingorani, and Juan~P. Casas.
\newblock The druggable genome and support for target identification and validation in drug development.
\newblock {\em Science Translational Medicine}, 9(383), March 2017.

\bibitem{XavierGlorotUnknown}
Xavier Glorot, Antoine Bordes, and Yoshua Bengio.
\newblock Deep sparse rectifier neural networks.
\newblock pages 315--323.

\bibitem{Goodfellow2014}
Ian~J. Goodfellow, Jean Pouget-Abadie, Mehdi Mirza, Bing Xu, David Warde-Farley, Sherjil Ozair, Aaron Courville, and Yoshua Bengio.
\newblock Generative adversarial networks.
\newblock 2014.

\bibitem{Grover2016}
Aditya Grover and Jure Leskovec.
\newblock node2vec: Scalable feature learning for networks.
\newblock 2016.

\bibitem{KarimollahHajian-TilakiUnknown}
Karimollah Hajian-Tilaki.
\newblock Receiver operating characteristic (roc) curve analysis for medical diagnostic test evaluation.
\newblock {\em Caspian Journal of Internal Medicine}, 4(2):627--.

\bibitem{Hamilton2017}
William~L. Hamilton, Rex Ying, and Jure Leskovec.
\newblock Inductive representation learning on large graphs.
\newblock 2017.

\bibitem{Hao_2016}
Tong Hao, Wei Peng, Qian Wang, Bin Wang, and Jinsheng Sun.
\newblock Reconstruction and application of protein–protein interaction network.
\newblock {\em International Journal of Molecular Sciences}, 17(6):907, jun 2016.

\bibitem{Hill_2008}
David~P Hill, Barry Smith, Monica~S McAndrews-Hill, and Judith~A Blake.
\newblock Gene ontology annotations: what they mean and where they come from.
\newblock {\em BMC Bioinformatics}, 9(S5), April 2008.

\bibitem{Huang_2021}
Zexi Huang, Arlei Silva, and Ambuj Singh.
\newblock A broader picture of random-walk based graph embedding.
\newblock In {\em Proceedings of the 27th ACM SIGKDD Conference on Knowledge Discovery \&; Data Mining}, KDD ’21. ACM, August 2021.

\bibitem{Hunter_2009}
S.~Hunter, R.~Apweiler, T.~K. Attwood, A.~Bairoch, A.~Bateman, et~al.
\newblock Interpro: the integrative protein signature database.
\newblock {\em Nucleic Acids Research}, 37(Database):D211–D215, jan 2009.

\bibitem{Hyndman_2011}
Rob~J. Hyndman.
\newblock {\em Moving Averages}, page 866–869.
\newblock Springer Berlin Heidelberg, 2011.

\bibitem{Jamali_2016}
Ali~Akbar Jamali, Reza Ferdousi, Saeed Razzaghi, Jiuyong Li, Reza Safdari, and Esmaeil Ebrahimie.
\newblock Drugminer: comparative analysis of machine learning algorithms for prediction of potential druggable proteins.
\newblock {\em Drug Discovery Today}, 21(5):718–724, may 2016.

\bibitem{Karczewski_2020}
Konrad~J. Karczewski, Laurent~C. Francioli, Grace Tiao, Beryl~B. Cummings, Jessica Alföldi, et~al.
\newblock The mutational constraint spectrum quantified from variation in 141,456 humans.
\newblock {\em Nature}, 581(7809):434–443, may 2020.

\bibitem{Kelleher_2022}
Keith~J Kelleher, Timothy~K Sheils, Stephen~L Mathias, Jeremy~J Yang, Vincent T Metzger, Vishal B Siramshetty, Dac-Trung Nguyen, Lars~Juhl Jensen, Dušica Vidović, Stephan C Schürer, Jayme Holmes, Karlie R Sharma, Ajay Pillai, Cristian G Bologa, Jeremy S Edwards, Ewy A Mathé, and Tudor I Oprea.
\newblock Pharos 2023: an integrated resource for the understudied human proteome.
\newblock {\em Nucleic Acids Research}, 51(D1):D1405–D1416, nov 2022.

\bibitem{Kim_2017}
Hae-Young Kim.
\newblock Statistical notes for clinical researchers: Chi-squared test and fisher’s exact test.
\newblock {\em Restorative Dentistry \&; Endodontics}, 42(2):152, 2017.

\bibitem{CanM.LeUnknown}
Can~M. Le.
\newblock Edge sampling using local network information.
\newblock {\em Journal of Machine Learning Research}, 22(88):1--29.

\bibitem{Myung_2013}
Jay~I. Myung, Daniel~R. Cavagnaro, and Mark~A. Pitt.
\newblock A tutorial on adaptive design optimization.
\newblock {\em Journal of Mathematical Psychology}, 57(3–4):53–67, jun 2013.

\bibitem{Qu2021}
Liang Qu, Huaisheng Zhu, Ruiqi Zheng, Yuhui Shi, and Hongzhi Yin.
\newblock Imgagn:imbalanced network embedding via generative adversarial graph networks.
\newblock 2021.

\bibitem{Raies_2022}
Arwa Raies, Ewa Tulodziecka, James Stainer, Lawrence Middleton, Ryan~S. Dhindsa, Pamela Hill, Ola Engkvist, Andrew~R. Harper, Slavé Petrovski, and Dimitrios Vitsios.
\newblock Drugnomeai is an ensemble machine-learning framework for predicting druggability of candidate drug targets.
\newblock {\em Communications Biology}, 5(1), nov 2022.

\bibitem{Rakshit_2023}
Gourav Rakshit, Komal, Pankaj Dagur, and Venkatesan Jayaprakash.
\newblock {\em Multi-Omics Approaches in Drug Discovery}, page 79–98.
\newblock Springer Nature Singapore, 2023.

\bibitem{Rasti_2018}
Saeid Rasti and Chrysafis Vogiatzis.
\newblock A survey of computational methods in protein–protein interaction networks.
\newblock {\em Annals of Operations Research}, 276(1–2):35–87, June 2018.

\bibitem{Reel_2021}
Parminder~S. Reel, Smarti Reel, Ewan Pearson, Emanuele Trucco, and Emily Jefferson.
\newblock Using machine learning approaches for multi-omics data analysis: A review.
\newblock {\em Biotechnology Advances}, 49:107739, July 2021.

\bibitem{Rokach}
Lior Rokach and Oded Maimon.
\newblock {\em Decision Trees}, page 165–192.
\newblock Springer-Verlag.

\bibitem{Schober_2018}
Patrick Schober, Christa Boer, and Lothar~A. Schwarte.
\newblock Correlation coefficients: Appropriate use and interpretation.
\newblock {\em Anesthesia \&; Analgesia}, 126(5):1763–1768, may 2018.

\bibitem{NitishSrivastavaUnknown}
Nitish Srivastava, Geoffrey Hinton, Alex Krizhevsky, Ilya Sutskever, and Ruslan Salakhutdinov.
\newblock Dropout: A simple way to prevent neural networks from overfitting.
\newblock {\em Journal of Machine Learning Research}, 15(56):1929--1958.

\bibitem{Tang_2015}
Jian Tang, Meng Qu, Mingzhe Wang, Ming Zhang, Jun Yan, and Qiaozhu Mei.
\newblock Line: Large-scale information network embedding.
\newblock In {\em Proceedings of the 24th International Conference on World Wide Web}, WWW ’15. International World Wide Web Conferences Steering Committee, may 2015.

\bibitem{Tsuji_2021}
Shingo Tsuji, Takeshi Hase, Ayako Yachie-Kinoshita, Taiko Nishino, Samik Ghosh, Masataka Kikuchi, Kazuro Shimokawa, Hiroyuki Aburatani, Hiroaki Kitano, and Hiroshi Tanaka.
\newblock Artificial intelligence-based computational framework for drug-target prioritization and inference of novel repositionable drugs for alzheimer’s disease.
\newblock {\em Alzheimer’s Research \&; Therapy}, 13(1), may 2021.

\bibitem{Wang_2019}
Jie Wang, Jiye Liang, Wenping Zheng, Xingwang Zhao, and Junfang Mu.
\newblock Protein complex detection algorithm based on multiple topological characteristics in ppi networks.
\newblock {\em Information Sciences}, 489:78–92, July 2019.

\bibitem{Yan_2007}
Shuicheng Yan, Dong Xu, Benyu Zhang, Hong-jiang Zhang, Qiang Yang, and Stephen Lin.
\newblock Graph embedding and extensions: A general framework for dimensionality reduction.
\newblock {\em IEEE Transactions on Pattern Analysis and Machine Intelligence}, 29(1):40–51, January 2007.

\bibitem{Yu_2022}
Lezheng Yu, Li~Xue, Fengjuan Liu, Yizhou Li, Runyu Jing, and Jiesi Luo.
\newblock The applications of deep learning algorithms on in silico druggable proteins identification.
\newblock {\em Journal of Advanced Research}, 41:219–231, November 2022.

\bibitem{Zhou2018}
Jie Zhou, Ganqu Cui, Shengding Hu, Zhengyan Zhang, Cheng Yang, Zhiyuan Liu, Lifeng Wang, Changcheng Li, and Maosong Sun.
\newblock Graph neural networks: A review of methods and applications.
\newblock 2018.

\end{thebibliography}

\end{document}